\documentclass{article}
\usepackage{ijcai25}

\usepackage{times}
\usepackage{soul}
\usepackage{url}
\usepackage[hidelinks]{hyperref}
\usepackage[utf8]{inputenc}
\usepackage[small]{caption}
\usepackage{graphicx}
\usepackage{amsmath}
\usepackage{amsthm}
\usepackage{booktabs}
\usepackage{algorithm}
\usepackage{algorithmic}
\usepackage[switch]{lineno}
\usepackage{amsfonts}
\usepackage{float}

\title{BRIGHT-VO: Brightness-Guided Hybrid Transformer for Visual Odometry with Multi-modality Refinement Module}

\author{
    Dongzhihan Wang$^1$
    \and
    Yang Yang$^2$\and Xuyang Chen$^3$\and Liang Xu $^*$
    \affiliations
    Shanghai University\\
    \emails
    \{wdzhihan, yyangy, sebastianchen, liang-xu\}@shu.edu.cn
    }

\begin{document}

\maketitle

\begin{abstract}
Visual odometry (VO) plays a crucial role in autonomous driving, robotic navigation, and other related tasks by estimating the position and orientation of a camera based on visual input. Significant progress has been made in data-driven VO methods, particularly those leveraging deep learning techniques to extract image features and estimate camera poses. However, these methods often struggle in low-light conditions because of the reduced visibility of features and the increased difficulty of matching keypoints. To address this limitation, we introduce BrightVO, a novel VO model based on Transformer architecture, which not only performs front-end visual feature extraction, but also incorporates a multi-modality refinement module in the back-end that integrates Inertial Measurement Unit (IMU) data. Using pose graph optimization, this module iteratively refines pose estimates to reduce errors and improve both accuracy and robustness. Furthermore, we create a synthetic low-light dataset, KiC4R, which includes a variety of lighting conditions to facilitate the training and evaluation of VO frameworks in challenging environments. Experimental results demonstrate that BrightVO achieves state-of-the-art performance on both the KiC4R dataset and the KITTI benchmarks. Specifically, it provides an average improvement of 20\% in pose estimation accuracy in normal outdoor environments and 25\% in low-light conditions, outperforming existing methods. This work is open-source at \url{https://github.com/Anastasiawd/BrightVO}.
\end{abstract}

\section{Introduction}
\label{sec:intro}
With the rapid advances in artificial intelligence technologies, the application domains of autonomous vehicles and mobile robots \cite{10466743,filipenko2018comparison} have broadened, encompassing a more diverse range of scenarios. 
As these systems increasingly operate in complex and dynamic environments, there is a growing emphasis on ensuring reliable localization, with a strong focus on improving the accuracy of pose estimation. Visual odometry (VO), a critical component in providing accurate and robust localization for autonomous systems, faces even greater challenges in such environments \cite{he2020review,gui2015review,aqel2016review}.
\begin{figure}[t]
    \centering
    \includegraphics[width=1\linewidth]{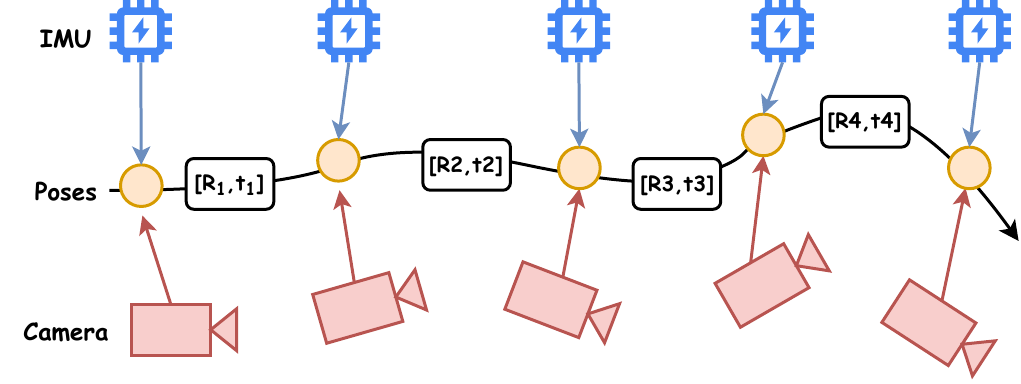}
    \caption{BrightVO estimates the motion of a camera using visual information, while also integrating IMU measurements to achieve more accurate pose estimation.}
    \label{fig:vo}
\end{figure}

VO is the task to track a vehicle or robot's pose, i.e., position and orientation over time \cite{mohamed2019survey}. Traditional localization techniques, such as GPS or LIDAR-based systems \cite{lin2022r,cai2019mobile}, often face limitations in urban canyons, indoor environments, or GPS-denied areas, where signal obstruction and degradation compromise their accuracy and reliability. In contrast, VO offers a solution to estimate the camera motion and localization, as it relies on visual sensors which are unaffected by such environmental constraints.
However, VO methods often struggle in low-light environments \cite{alismail2016direct,agostinho2022practical}. These environments, commonly encountered during nighttime or in areas with limited lighting, present difficulties for VO algorithms. Low-light conditions introduce issues such as poor feature visibility, increased sensor noise, and reduced contrast, all of which significantly affect the accuracy and robustness of traditional VO approaches. As a result, many existing systems are unable to maintain reliable localization or trajectory estimation when operating under such conditions.

To address these challenges, we propose BrightVO, a novel VO framework specifically designed to operate effectively in low-light environments shown in Figure \ref{fig:vo}. BrightVO uses a Transformer-based architecture \cite{vaswani2017attention}, a cutting-edge approach in deep learning that excels in modeling long-range dependencies and complex feature relationships within images. 
The key advantage of Transformer models lies in their self-attention mechanism, which allows the model to focus selectively on the most relevant features of an image. Additionally, we introduce a brightness estimation module, which uses convolutional layers to extract brightness features from the image. This allows the Transformer to focus on the illumination information, thereby addressing the limitations of traditional methods that struggle to effectively extract features due to low image quality in low-light conditions.
Furthermore, BrightVO's ability to integrate multi-modality data, such as Inertial Measurement Unit (IMU), further enhances its robustness. IMU provides complementary information to visual features, particularly in scenarios where visual input is noisy or sparse \cite{qin2018vins,huai2022robocentric}. By incorporating this measurement into a back-end based on graph optimization, which minimizes the cumulative drift and maintain consistency through iterative corrections, the model refines its pose estimates, improving accuracy over long sequences in challenging light conditions. 

The main contributions of the paper can be summarized as follows:
\begin{enumerate}
    \item We propose a brightness-guided Visual Transformer (ViT) \cite{dosovitskiy2020image}, which can learn the relative camera pose from multi-modality inputs through end-to-end training.
    \item  We design a back-end refinement block, using graph optimization with IMU inputs to iteratively improve the pose estimation accuracy. Experiments show that we achieve an average improvement of $20\%$ pose estimation accuracy in normal outdoor scenes and $25\%$ in low-light conditions compared to other methods.
    \item We create a low-light scene dataset using the CARLA \cite{dosovitskiy2017carla} simulator, which can be used for training and evaluating various VO frameworks.
\end{enumerate}

In the rest of this paper, Section 2 reviews related work,  Section 3 details the architecture of BrightVO, Section 4 presents the experimental setup and results. Finally, Section 5 concludes the paper.
\section{Related Work}

\subsection{Monocular visual odometry}
Monocular visual odometry is a technique used to estimate the motion trajectory of a camera in a 3D space from a sequence of images captured by a single camera. It relies on computer vision algorithms to calculate the camera's pose by analyzing changes between consecutive image frames. Currently, there are two main approaches to monocular visual odometry: traditional geometry-based methods and deep learning-based methods.

Geometry-based methods such as ORB-SLAM3 \cite{campos2021orb} extract key feature points from images and perform feature matching between consecutive frames, subsequently estimating camera motion based on geometric relationships. In contrast, methods such as DSO \cite{wang2017stereo} and LSD-SLAM \cite{engel2014lsd} do not rely on feature points; instead, they directly utilize pixel intensity differences to estimate camera motion. These approaches optimize camera poses by minimizing the photometric error between adjacent frames. However, due to the presence of noise and errors, the accumulated drift may increase over time, potentially leading to progressively inaccurate pose estimations.

In recent years, deep learning-based  methods such as DeepVO \cite{wang2017deepvo} and TartanVO \cite{wang2021tartanvo} have utilized deep neural networks to extract more robust features. These methods typically employ an end-to-end approach, training neural network models to directly estimate the relative pose of the camera from input images or image pairs. However they also exhibit greater adaptability to challenges such as challenging lighting conditions.

\subsection{VO under low-light condition}
Challenging lighting conditions present significant challenges to VO. Existing methods for these conditions adopt image enhancement algorithms prior to improve image rightness and enrich image detail features, thereby enhancing the accuracy of VO.
Light-SLAM \cite{zhao2024light} replaces traditional handcrafted features with LightGlue \cite{sarlin2020superglue} network to improve feature extraction in dark environments \cite{burri2016euroc}.  \cite{zhang2018visual} combines VO with 3D point cloud technology to enhance scene understanding under low-light conditions.
However, these approaches still rely heavily on local feature matching, which may not fully capture long-range dependencies and complex contextual relationships in images. Also, Light-SLAM has not been open-sourced, limiting reproducibility and preventing us from directly evaluating its performance.
\subsection{Transformers-based VO}
\begin{figure*}[t!]
    \centering
    \includegraphics[width=1\linewidth]{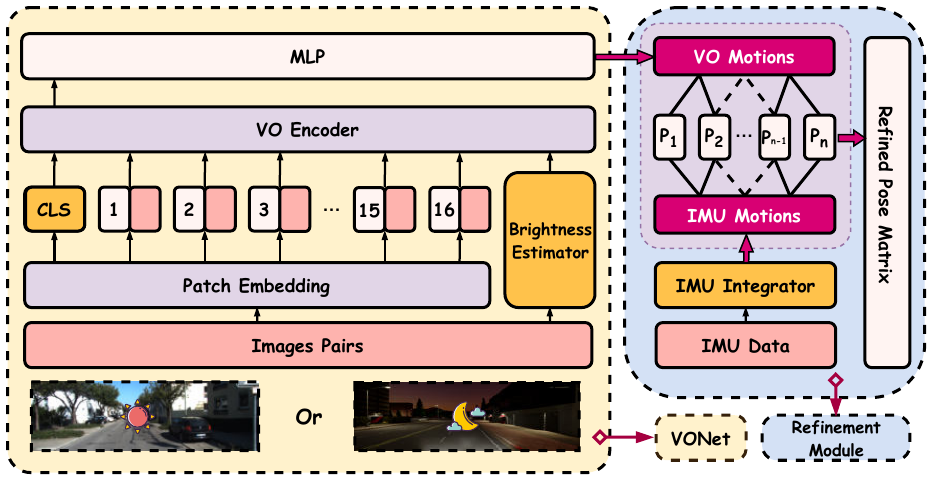}
    \caption{The pipeline of BrightVO, which begins with the input of consecutive image frames along with other modality information. In the VONet, images are transformed into vectors and passed into the Transformer Encoder after embedding. The global features are stored in the \( \texttt{cls\_token} \), which is then mapped to a 6-dimension vector via an MLP. This  vector represents the output initial VO motion. In the Refinement Module, IMU measurements are integrated to motions in an integrator. The VO motion, along with IMU motion, is fed into the PGO module, where iterative optimization occurs, yielding a high-precision pose estimate.
}
    \label{fig:pipeline}
\end{figure*}
In recent years, Transformer models have shown exceptional performance in both natural language processing and computer vision fields. Consequently, end-to-end VO methods can leverage the self-attention mechanism of Transformers to effectively capture global features in images, enhancing the temporal information processing in complex scenes. \cite{han2020survey,wu2020visual}.

TSformer-VO \cite{franccani2023transformer} proposes an end-to-end Transformer-based architecture for estimating 6 degrees of freedom (DoF) camera poses. Based on the TimeSformer \cite{bertasius2021space} model, it extracts features from image sequences through both spatial and temporal self-attention mechanisms. However, this method achieves reduced accuracy on the KITTI \cite{geiger2012we} dataset. \cite{memmel2023modality} propose a Transformer model which is a modality-agnostic. Experimental results indicate the model achieves robust performance in indoor navigation tasks.

These models successfully integrate Transformer technology into VO and demonstrate superior results on common datasets compared to traditional methods. However, these models were not specifically designed to address the challenges posed by low-light conditions. Additionally, existing approaches cannot incorporate traditional mathematical optimization methods \cite{martinez2022ransac,carlone2015initialization}, potentially resulting in noticeable scale drift and a lack of effective correction in extended sequences.

\section{Approaches}

\subsection{Preliminaries} The core task of visual odometry (VO) is to estimate the camera's position and pose changes. From an End-to-End (E2E) perspective, VO is data-driven and aims to automate the entire process of motion estimation from image input to output using neural networks. The input of Visual Odometry (VO) consists of a sequence of RGB images \( I_t \) and \( I_{t+1} \) captured by a monocular camera, along with other modality information such as IMU sensor data and depth maps. The output is a 6-DOF vector that includes both the translational vector \( \mathbf{t} \) and the rotational matrix \( \mathbf{R} \) information. The transformation from time \( t \) to time \( t+1 \) can be expressed as:
\begin{equation}
\mathbf{T}_{t \rightarrow t+1} = \begin{bmatrix} \mathbf{R}_t & \mathbf{t}_t \end{bmatrix},
\end{equation}
where \( \mathbf{R} \in \mathbb{R}^{3 \times 3} \) is the rotation matrix representing the rotation change of the camera from time \( t \) to \( t+1 \),
\( \mathbf{t}_t \in \mathbb{R}^3 \) is the translation vector representing the translation change of the camera from time \( t \) to \( t+1 \).
Thus, the entire process can be mathematically represented as:

\begin{equation}
 f(I_t, I_{t+1}) \rightarrow \mathbf{T}_{t \rightarrow t+1} = \begin{bmatrix} \mathbf{R}_t & \mathbf{t}_t \end{bmatrix}.
\end{equation}   

The goal of our model is to extract deep features from the input multi-modality information, solve for the camera's motion, and achieve superior performance in pose estimation under low-light conditions.
\subsection{Model Architecture}
Our pipeline consists of a front-end VOnet and a back-end refinement module, as shown in the Figure \ref{fig:pipeline}. The front-end VOnet is designed based on a basic Encoder-Decoder architecture. The encoder uses an enhanced ViT model to encode multi-modality input information and we propose a brightness-guided strategy in this part. The decoder is relatively simple, consisting of a multi-layer perceptron (MLP) that progressively extracts features and generates the final 6-DOF pose vector. The back-end refinement block uses Pose Graph Optimization to iteratively refine the pose estimation for more accurate results.

\textbf{Transformer Encoder:}
The encoder is responsible for extracting features from the input image sequence and passing these features to the decoder. First, the input images are resized to 224 \(\times\) 224 pixels, a size consistent with the pre-trained model. Our experiments demonstrate that this resizing does not significantly impact the model's performance. The images are then divided into several 16 \(\times\) 16 patches and passed through a convolutional layer, which transforms them into fixed-dimensional vectors. To preserve spatial information, the encoder adds positional encoding to each patch. 
\begin{figure}[t]
    \centering
    \includegraphics[width=0.85\linewidth]{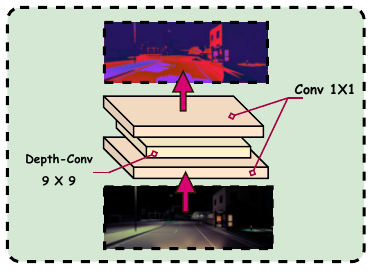}
    \caption{The overview of Brightness Estimator. The image at the bottom is the input. After passing through a network composed of two 1x1 convolutions and a 9x9 depth-wise convolution, we obtain the brightness feature map shown in the top image, where different RGB values represent different illumination intensities.}
    \label{fig:estimator}
\end{figure}

Next, a brightness estimator is used to extract brightness features from the image as illustrated in Figure \ref{fig:estimator}, which are then combined with the image features and input into the Transformer model. This module is particularly effective at handling lighting variations. A similar module is also used in Retinexformer \cite{cai2023retinexformer} to recover a normally exposed image from a low-exposure image using a UNet-based architecture \cite{ronneberger2015u}. However, we believe that it is unnecessary to up-sample the image to a normally exposed version in the VO task as perceived by the human eye. Therefore, we simplify the module as follows to enhance the performance of pose estimation.
\begin{equation}
(\mathbf{I}_{\text{br}}, \mathbf{F}_{\text{br}}) = \mathcal{E}(\mathbf{I}, \mathbf{L}_p),
\end{equation}
where \( \mathcal{E} \) denotes the brightness estimator and \( \mathbf{I} \) represents the resized images. \( \mathcal{E} \) takes \( \mathbf{I} \) and its brightness prior map \( \mathbf{L}_p \in \mathbb{R}^{H \times W} \) as inputs. \( \mathbf{L}_p = \text{mean}_c(\mathbf{I}) \), \(\text{mean}_c\) indicates the operation that calculates the mean values for each pixel along the channel dimension. \(\mathcal{E}\) outputs the brightness-enhanced image \( \mathbf{I}_{\text{br}} \) and the brightness feature \( \mathbf{F}_{\text{br}} \in \mathbb{R}^{H \times W \times C} \).

\begin{figure}[t]
    \centering
    \includegraphics[width=0.85\linewidth]{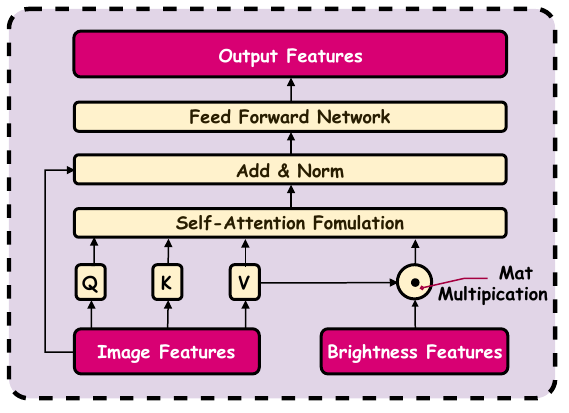}
    \caption{ The overview pf Transformer Layer in the VO Encoder. After patch embedding, image features are used to generate Q (queries), K (keys), and V (values). These, combined with brightness information, are employed to compute attention scores. The self-attention mechanism is then applied, and the resulting hidden states are passed through a feed-forward network to produce the final output features.   }
    \label{fig:attention}
\end{figure}

The encoder consists of several Transformer layers, as shown in the Figure \ref{fig:attention}, each containing a self-attention mechanism and a feed-forward network to capture complex features in the image. The final output of the encoder is a hidden state that contains both image and brightness features, which is used for subsequent pose estimation.

We also reshape \( \mathbf{F}_{\text{br}} \) into \( \mathcal{V} \in \mathbb{R}^{HW \times C} \)  and then the self-attention is formulated as:
\begin{equation}
Atten(Q, K, V, \mathcal{V}) = (V \odot \mathcal{V}) \, \mathop{softmax}(\frac{K^T Q }{\alpha}),
\end{equation}
where \(\alpha \in \mathbb{R}^1\) is a learnable parameter that adaptively scales the matrix multiplication. $\mathop{Atten}$ represents the self-attention mechanism.

As value ($V$) contains the actual feature information and represents the real content or information of each position in the feature space, this self-attention mechanism allows the model to weigh the importance of different parts of the input sequence, effectively capturing the relationships between patches and encoding complex image features.

\textbf{Transformer Decoder:} The decoder is designed to process the features extracted by the encoder and produce the final pose estimation. It follows a simple structure leveraging a multi-layer perceptron (MLP) and normalization techniques to refine the feature representation. The mathematical formulation of the decoder can be summarized as follows:
\begin{align}
    x'&= \mathop{DropPath}(\mathop{MLP}(\mathop{LayerNorm}(x))), \\
    \hat{T} &= W_{out}\cdot x' + b_{out},
\end{align}
where $W_{out} \in \mathbb{R}^{768 \times 6}$ is the weight matrix of the output layer, which transforms the feature vector $x'$ into the final output pose vector. $b_{out} \in \mathbb{R}^{6}$ is the bias vector of the output layer, added after the matrix multiplication to provide additional flexibility. $\hat{T} \in \mathbb{R}^{6}$ is the final output, which is a 6-DOF pose vector including rotational and translational components.

\textbf{Back-end refinement module:} The refinement module in our approach is based on Pose Graph Optimization (PGO) \cite{fu2024islam}, where the goal is to optimize the trajectory estimates by minimizing errors across multiple sensor modalities. This back-end module enables two modalities with different error characteristics VO and IMU to mutually verify and correct each other through graph constraints, leading to a more precise and robust trajectory estimation.

In our approach, we reshape the pose estimated by VOnet into the SE(3) form to adapt to the optimization mechanism of  PGO. The motion of the IMU is integrated from raw sensor data using an IMU integrator. Specifically, the IMU’s motion is calculated by integrating the acceleration measurements, as well as by considering the positional estimates from Global Navigation Satellite System (GNSS) when available. The equations governing the motion of the IMU are as follows:
\begin{align}
   \Delta \mathbf{R}_{ik+1} &= \Delta \mathbf{R}_{ik} \exp(\mathbf{w}_k \Delta t), \\
    \Delta \mathbf{v}_{ik+1} &= \Delta \mathbf{v}_{ik} + \Delta \mathbf{R}_{ik} \mathbf{a}_k \Delta t, \\
    \Delta \mathbf{p}_{ik+1} &= \Delta \mathbf{p}_{ik} + \Delta \mathbf{v}_{ik} \Delta t + \frac{1}{2} \Delta \mathbf{R}_{ik} \mathbf{a}_k \Delta t^2,
\end{align}
where $\Delta \mathbf{R}_{ik}$ is the pre-integrated rotation between the $i$-th and $k$-th time steps. $\Delta \mathbf{v}_{ik}$ is the pre-integrated velocity between the $i$-th and $k$-th time steps. $\Delta \mathbf{p}_{ik}$ is the pre-integrated position between the $i$-th and $k$-th time steps. $\mathbf{a}_k$ is the linear acceleration at the $k$-th time step. $\mathbf{w}_k$ is the angular velocity at the $k$-th time step. $\Delta t$ is the time interval between the $k$-th and $(k+1)$-th time steps.

The last term, we also use GNSS data at each time step as correction. This correction ensures the position stays globally aligned with the real-world coordinates.  Therefore, the final error of the VO motions $\mathbf{T}_{ij}$ and IMU poses $\mathbf{p}_{IMU}$ can be defined as the weighted summation of the two constraints:
\begin{small}
    \begin{equation}
        \mathcal{L} = \sum_{(i,j) \in \mathcal{E}} \left\| \mathbf{p}_{ij} - \mathbf{T}_{ij} \right\|^2_{\Sigma_{ij}} + \lambda\sum_{(i,j) \in \mathcal{E}} \left\| \mathbf{p}_{ij}- \mathbf{p}_{\mathop{IMU}} \right\|^2_{\Sigma_{ij}},
    \end{equation}
\end{small}
where $\mathcal{E}$ represents all image frames and $\mathbf{p}_{ij}$ is the relative pose between frames $i$ and $j$, which is defined as a parameter and iteratively optimized during the PGO.

We then employ a Levenberg-Marquardt (LM) algorithm in PyPose \cite{wang2023pypose} to solve the PGO process, which leads to more accurate and consistent trajectory estimation and ensures robustness across a variety of scenarios, especially when GNSS data is available to correct for long-term drift.
\section{Experiments}

\subsection{Datasets preparation}
\textbf{KiC4R:} To validate the performance of our proposed model under low-light conditions, we need a sufficiently large dataset that includes various lighting conditions for training and testing. However, to date, we have not been able to find a suitable dataset that meets these criteria. The low-light scenes in the TUM dataset \cite{keimel2012tum} are limited to indoor environments, and TartanAir \cite{wang2020tartanair} only provides the "\texttt{abandonedfactory\_night}" sequence as a low-light sequence, which is insufficient for training purposes. Many previous works \cite{rashed2019fusemodnet} have addressed the lack of such datasets by simulating nighttime scenes using networks or capturing real-world data. However, we believe these methods are not the most efficient or comprehensive ways to obtain the necessary data. 

\begin{figure}[h]
     \centering
     \includegraphics[width=1\linewidth]{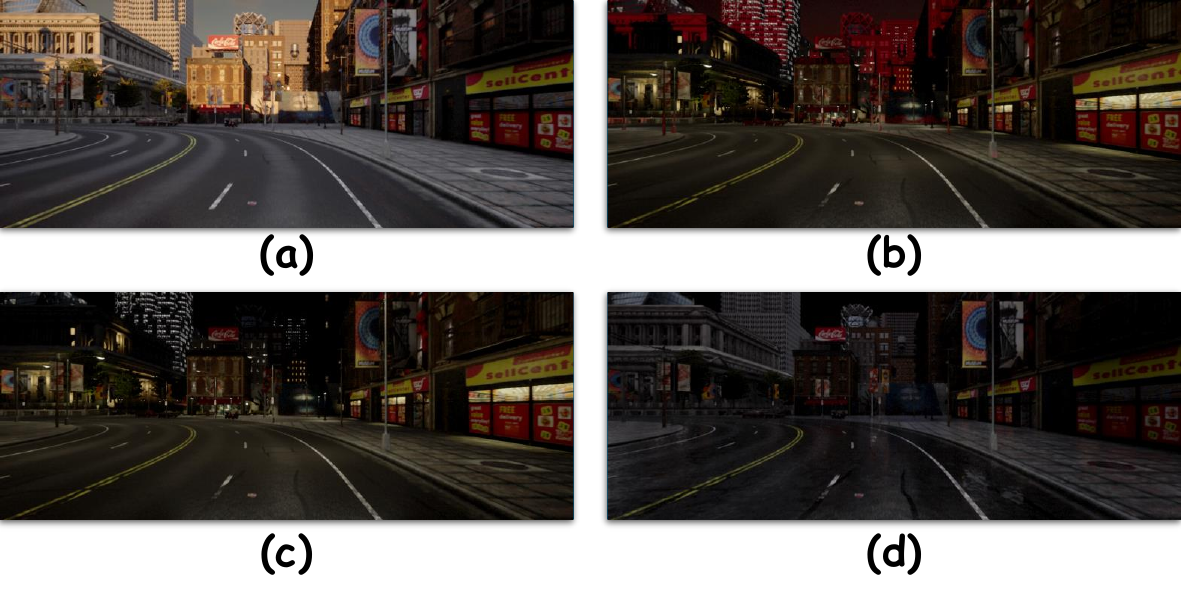}
     \caption{The overview of 4 light conditions in KiC4R. (a) Dusk. (b) Nighttime. (c) Mid-night. (d) Extreme weather.}
     \label{fig:light_conditions}
 \end{figure}

To address this issue, we created a dataset called KiC4R using the CARLA simulator. The KiC4R dataset includes seven sequences i.e., 00-06 and features four types of low-light conditions: dusk, night, midnight, and extreme weather shown in Figure \ref{fig:light_conditions}. We model the passage of time by adjusting the sun's direct angle in the simulator. When the angle is negative, the scene transitions to nighttime, with the time progressively changing, eventually reaching midnight. Additionally, vehicle headlights and streetlights are activated to simulate realistic nighttime street scenes. To emulate low-light conditions induced by extreme weather, we generated two sequences i.e., 00 and 03 by manipulating the cumulus cloud cover, precipitation, and road water accumulation, thereby simulating low-light scenarios under heavy rainfall.

SHIFT \cite{sun2022shift} also provides a large-scale, multi-scenario, and multi-task dataset using CARLA . However, due to differences in data annotation formats, this dataset cannot be directly used for VO task training and evaluation, which posed some challenges for experimentation. For ease of comparison with other prominent works, KiC4R consists of both RGB and IMU sensor data and adopts the same annotation format as KITTI.

We ran the CARLA simulator on a Windows PC equipped with an NVIDIA RTX 4090 GPU. Sensor data was collected using CarlaScenes and the official CARLA Python API. To generate sensor data from the simulator, we selected maps 01-07 and 10 for urban scenes similar to those found in datasets such as \textit{Virtual KITTI} \cite{cabon2020virtual}.

We ultimately collected over 37,000 RGB images with a resolution of 512 $\times$ 1392 using the approaches described above. The corresponding maps, lighting conditions, and sequence lengths for these sequences are shown in the table \ref{tab:sequence}.
\begin{table}[h]
  \centering
  \begin{tabular}{@{}cccc@{}}
    \toprule
    Sequence & Map & Light Condition & Lengths\\
    \midrule
    00 & Town\_01 & E & 5000\\
    01 & Town\_01 & D & 3200 \\
    02 & Town\_02 & M & 4000 \\
    03 & Town\_03 & E & 8000 \\
    04 & Town\_06 & D & 7446 \\
    05 & Town\_07 & N & 8000 \\
    06 & Town\_10 & N & 2000 \\
    \bottomrule
  \end{tabular}
  \caption{Details of KiC4R sequences, where D, N, M, E represents of dusk, nighttime, mid-night and extreme weather and Lengths represent how many frames in the sequence.}
  \label{tab:sequence}
\end{table}

\subsection{Experiment setup}
\begin{table*}[t]
  \centering
  \begin{tabular}{@{}c|cccccccccccc@{}}
    \toprule
     Sequence&00 & 01 & 02 & 04 & 05 & 06 & 07 & 08 & 09 & 10&Avg \\
    \midrule
    ORB-SLAM2 & \textbf{1.3}&10.4&5.7&\textbf{0.2}&\textbf{0.8}&\textbf{0.8}&\textbf{0.5}&3.6&3.2&\textbf{1.0}&2.75 \\
    ORB-SLAM3 & 6.77&-&30.50&0.93&5.54&16.61&9.70&60.69&7.90&8.65&-  \\
    \midrule
    DeepVO &95.92&68.26&150.56&5.65&54.86&88.47&7.96&68.19&30.70&22.76&59.33   \\
    DPVO &111.97&12.69&123.40&0.68&58.96&54.78&19.26&115.90&75.10&13.63&58.64\\
    DPV-SLAM &8.30&11.86&39.64&0.78&5.74&11.6&1.52&110.9&76.7&13.7&28.09   \\
    \midrule
    TSFormer-VO &46.52&160.55&55.24&3.06&61.38&88.31&31.49&26.46&23.68&22.70&51.93\\
    \textbf{Ours} &2.12&\textbf{2.36}&\textbf{2.52}&0.44&2.31&2.7&2.04&\textbf{2.92}&\textbf{2.26}&2.11&\textbf{2.18} \\
    \bottomrule
  \end{tabular}
  \caption{Absolute Trajectory Error (ATE) on 10 sequences on \textit{KITTI} dataset, given in meters. Due to the absence of raw IMU data in sequence 3, we had to discard this sequence. ORB-SLAM 2, 3 are feature-based methods; DeepVO, TartanVO, DPVO and DPV-SLAM are traditional learning-based methods; TSFormer-VO and ours are transformer-based methods.}
  \label{tab:kitti_ate}
\end{table*}

\begin{table*}[t]
  \centering
  \begin{tabular}{@{}c|cccccccccc@{}}
    \toprule
    Sequence & \multicolumn{2}{c}{06} & \multicolumn{2}{c}{07} & \multicolumn{2}{c}{09} & \multicolumn{2}{c}{10} & \multicolumn{2}{c}{Avg}\\
    
     &$t_{rel}$ & $r_{rel}$ &$t_{rel}$ & $r_{rel}$ & $t_{rel}$ & $r_{rel}$ & $t_{rel}$ & $r_{rel}$ & $t_{rel}$ & $r_{rel}$\\ 
    \midrule
    TartanVO &4.72& 2.95&4.32 &3.41&6.00 &3.11&6.89 &2.73&5.48& 3.05 \\
    DPV-SLAM &4.95 &\textbf{0.16}&1.29 &\textbf{0.24}&17.69 &\textbf{0.23}&6.32 &\textbf{0.23}&7.56 &\textbf{0.22}\\
    \textbf{Ours} &\textbf{1.35}& 0.98&\textbf{1.00}& 1.22&\textbf{1.31} &0.76&\textbf{1.27}& 0.99&\textbf{1.23}&0.99\\
    \bottomrule
  \end{tabular}
  \caption{Sence TartanVO only reports results for sequences 06,07,09,10 in relative pose error (RPE), here we compared our method with two advanced approaches using the $t_{rel}$/$r_{rel}$ metrics.}
  \label{tab:kitti_rpe}
\end{table*}

\begin{table*}[h!]
  \centering
  \begin{tabular}{@{}c|cccccccccc@{}}
    \toprule
    Sequence & \multicolumn{2}{c}{00} & \multicolumn{2}{c}{01} & \multicolumn{2}{c}{02} & \multicolumn{2}{c}{03} & \multicolumn{2}{c}{Avg}\\
    
     &$t_{rel}$ & $r_{rel}$ &$t_{rel}$ & $r_{rel}$ & $t_{rel}$ & $r_{rel}$ & $t_{rel}$ & $r_{rel}$ & $t_{rel}$ & $r_{rel}$\\ 
    \midrule
    ORB-SLAM3 &1.11& 1.15&1.08 &0.99&1.31 &1.21&1.44 &1.20& 1.24 & 1.14   \\
    TartanVO &5.62& 2.44&5.32 &3.21&4.33 &3.72&3.78 &1.88& 4.76 & 2.81   \\
    DPVO &1.29 &0.97&1.32 &0.98&1.28&0.89&1.31 &\textbf{0.75}& 1.30 & 0.90\\
    \textbf{Ours} &\textbf{0.74} & \textbf{0.93} &\textbf{0.62}& \textbf{0.85}&\textbf{0.13} & \textbf{0.13} &\textbf{0.93}& 0.94&\textbf{0.61} &\textbf{0.71} \\
    \bottomrule
  \end{tabular}
  \caption{RPE on 4  sequences on KiC4R. Here we compared our method with ORB-SLAM3, TartanVO and DPVO using the $t_{rel}$/$r_{rel}$ metrics.}
  \label{tab:kic4r_rpe}
\end{table*}

To evaluate the effectiveness of our proposed method, we conducted experiments on the KITTI and KiC4R datasets. We compared our approach with several state-of-the-art frameworks, including ORB-SLAM2 \cite{mur2017orb}, ORB-SLAM3, DeepVO , TartanVO, DPVO \cite{teed2024deep}, and DPV-SLAM \cite{lipson2025deep}. In our experiments, only the front-end VONet was trained. Specifically, we used sequences 01, 03, 07, and 08 from the KITTI dataset and sequences 04-06 from the KiC4R dataset as the training set. The inputs contain $N$ monocular image pairs with resolution 512×1392. The outputs are $N$ poses in SE(3). For evaluation, we adpoted two widely used metrics: Absolute Trajectory Error (ATE) and Relative Pose Error (RPE) \cite{prokhorov2019measuring}. Shown as follows:
\begin{align}
    \mathop{ATE} &= \sqrt{\frac{1}{N} \sum_{i=1}^{N} \| \mathbf{p}_i - \hat{\mathbf{p}}_i \|^2}, \\
    \mathop{RPE} &= \sqrt{\frac{1}{N-1} \sum_{i=1}^{N-1} \| (\mathbf{T}_{i}^{-1} \mathbf{T}_{i+1}) - (\hat{\mathbf{T}}_{i}^{-1} \hat{\mathbf{T}}_{i+1}) \|^2},
\end{align}
where $\mathbf{p}$ is the camera pose of each frame, $i$ is the number of frame, $\mathbf{T}$ and $\hat{\mathbf{T}}$ represent the estimated and ground-truth translation vector.

To deal with the large data requirements of ViTs, we use the pre-trained model \texttt{"vit-base-patch16-224"} made publicly available  by \cite{dosovitskiy2020image}. We then trained and finetuned our model for 250 epochs on a single NVIDIA RTX 4090 GPU with a batch size of 12. The weights of the model are updated using \texttt{AdamW} optimizor with a learning rate of $1 * 10^{-4}$.

\subsection{Experiment results}

\begin{figure*}[t]
    \centering
    \includegraphics[width=1\linewidth]{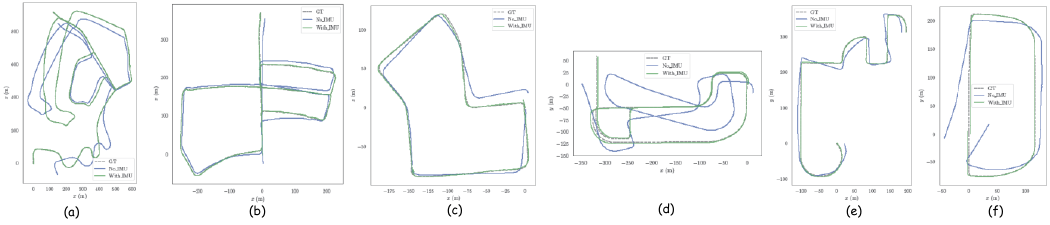}
    \caption{The overview of trajectory plots with/without IMU measurements. (a)-(c) represent sequence 02, 05, 07 on KITTI dataset. (d)-(f) represent sequence 00-03 on KiC4R dataset.}
    \label{fig:ablation}
\end{figure*}

 We first conducted experiments on the KITTI dataset to validate the performance of our VO network under normal lighting conditions. The model was trained for 40 hours using sequences 01, 03, 07, and 08. Following this, we evaluated the model across all sequences, utilizing the evo tool to align the results, recover the scale, and compute the ATE and RPE. These results were then compared against several state-of-the-art methods, including ORB-SLAM2, ORB-SLAM3, TartanVO, DPVO, DPV-SLAM, and TSFormer-VO. Notably, DPVO represents an improvement over DROID-SLAM \cite{teed2021droid}, which is widely recognized as a state-of-the-art method. The results are presented in Table \ref{tab:kitti_ate} and \ref{tab:kitti_rpe}.

The experiment results demonstrate that while ORB-SLAM2 achieves minimal ATE in certain sequences (e.g., 04-07), BrightVO consistently outperforms it, achieving a 20\% improvement in average ATE over all 10 sequences. In comparison to methods lacking backend optimization, such as TartanVO, DeepVO, and TSFormer-VO, BrightVO demonstrates a substantial reduction in error nearly 96\%. Furthermore, in comparison to DPV-SLAM, which also includes mapping process, BrightVO achieves superior performance, benefiting from its multi-modalaity refinement approach. These results collectively underscore that our model not only functions effectively under normal lighting conditions but also delivers state-of-the-art pose estimation accuracy.

We then conduct experiments on the KiC4R dataset to assess whether our method enhances VO performance in low-light scenarios. The test set comprises four sequences i.e., 00-03 from the KiC4R dataset, containing a total of 20,200 images after excluding frames affected by camera shake. For evaluation, we computed the RPE on every 100 meters for each sequence and compared our results with TartanVO, ORB-SLAM3, and DPVO. Our experimental results demonstrate that BrightVO consistently achieves smaller relative errors across all four sequences, outperforming existing approaches as shown in Table \ref{tab:kic4r_rpe}. Our method decreases the average relative error by approximately 50\% compared with state-of-the-art methods. We attribute this significant improvement to the use of longer sequences and more extreme lighting conditions, where BrightVO benefits from the robust feature extraction capabilities of the Brightness-Guided ViT for long sequences, coupled with back-end optimization that minimizes drift during extended operations.

\subsection{Ablation study}
\textbf{Modality independent:}
In real-world conditions, IMU and GNSS measurments may fail when GNSS signals are obstructed or completely lost, which prevents GNSS from providing accurate positional corrections. Similarly, IMU data can be compromised due to factors like high vibrations, rapid accelerations, or sensor malfunctions, leading to inaccuracies or even complete data loss. Since the KiC4R dataset does not include GNSS data for correction, the experiments conducted in the previous section have already demonstrated that BrightVO can still achieve robust estimation accuracy in the absence of GNSS corrections. To further assess BrightVO’s performance without IMU data, we removed the refinement module while keeping all other settings unchanged, and conducted experiments on the KITTI and KiC4R datasets. The results in Figure \ref{fig:ablation} show that, in the absence of the back-end refinement module, BrightVO experienced significant drift in both normal and low-light conditions. This highlights the critical role of the refinement module in ensuring the stability and accuracy of BrightVO.
\begin{figure}[t]
    \centering
    \includegraphics[width=1\linewidth]{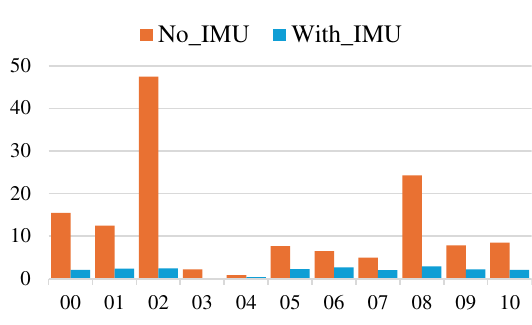}
    \caption{The illustration of ATE on KITTI sequences with/without IMU measurements. Due to the absence of IMU raw data on sequence 03, the figure only contains result without IMU inputs. }
    \label{fig:wo_imu}
\end{figure}
In addition, as shown in Figure \ref{fig:wo_imu}, in shorter sequences, such as 03, 04, and 07, even when the back-end refinement module is removed, BrightVO demonstrates minimal estimation errors. This suggests that our model is capable of maintaining reliable estimation accuracy, even in short-term absence of IMU measurements.

\section{Conclusion}
In this paper, we propose BrightVO, a model designed to enhance accuracy in VO under low-light conditions. Our model consists of a ViT-based VO network and an optimization module that integrates multi-modality information. By combining visual information with IMU data, BrightVO significantly improves pose estimation accuracy under extreme lighting conditions.

We conducted extensive experiments on the KiC4R and KITTI datasets and the experimental results demonstrate that BrightVO outperforms existing VO methods, achieving state-of-the-art performance in various environments. We also designed several ablation experiments which confirmed that BrightVO still maintain good estimation accuracy even with short-term IMU data loss.

Ultimately, BrightVO not only excels in low-light scenarios but also operates stably under normal environmental conditions, demonstrating broad application potential. Our research provides a new perspective for data-driven VO tasks and offers strong support for future applications in autonomous driving, robotics, and other fields.
\section*{Acknowledgments}
This work is supported by the National Natural Science Foundation of China under grant 62373239. Specially, we would like to thank Dr. Chen Wang for their technical support.
\bibliographystyle{named}
\bibliography{ijcai25}

\end{document}